\documentclass[a4paper,10pt,twoside]{article}
\usepackage{clin}        % Stylefile for CLIN Journal
\usepackage{harvard}     % Bibliography Stylefile
\usepackage{booktabs}
\usepackage{graphicx}
\usepackage{hyperref}
\usepackage{todonotes}
\usepackage[T1]{fontenc}

\usepackage{csquotes}

\begin{document}

\title{RobBERTje: A Distilled Dutch BERT Model}

\author{Pieter Delobelle$^*$ \email{pieter.delobelle@cs.kuleuven.be}\\
{\normalsize \bf Thomas Winters}$^*$ \email{thomas.winters@cs.kuleuven.be}\\
{\normalsize \bf Bettina Berendt}$^{**}$ \email{bettina.berendt@cs.kuleuven.be}\\
\AND \addr{$^*$Department of Computer Science; Leuven.AI, KU Leuven, Belgium}
\AND \addr{$^{**}$TU Berlin and Weizenbaum Institute, Germany; Leuven.AI and KU Leuven, Belgium}}

\maketitle\thispagestyle{empty} % extra pagestyle command for first page

%To be filled in by the editors
%Please leave commented out
%\jmlrheading{vol}{year}{pages}{Submission date}{Publication date}{authors}
%\copyright

\begin{abstract}
Pre-trained large-scale language models such as BERT have gained a lot of attention thanks to their outstanding performance on a wide range of natural language tasks.
However, due to their large number of parameters, they are resource-intensive both to deploy and to fine-tune.
Researchers have created several methods for distilling language models into smaller ones to increase efficiency, with a small performance trade-off.
In this paper, we create several different distilled versions of the state-of-the-art Dutch RobBERT model and call them RobBERTje.
The distillations differ in their distillation corpus, namely whether or not they are shuffled and whether they are merged with subsequent sentences.
We found that the performance of the models using the shuffled versus non-shuffled datasets is similar for most tasks and that randomly merging subsequent sentences in a corpus creates models that train faster and perform better on tasks with long sequences.
Upon comparing distillation architectures, we found that the larger DistilBERT architecture worked significantly better than the Bort hyperparametrization.
Interestingly, we also found that the distilled models exhibit less gender-stereotypical bias than its teacher model.
Since smaller architectures decrease the time to fine-tune, these models allow for more efficient training and more lightweight deployment of many Dutch downstream language tasks.
\end{abstract}

\section{Introduction}

Large-scale pre-trained language models such as BERT \cite{devlin2019bert} have revolutionized many natural language processing tasks thanks to their outstanding performance on downstream tasks.
Initially, a BERT model is pre-trained on a large corpus of text sequences to predict which words---or more precisely tokens---are likely on masked positions in a sentence. 
This task, called Masked Language Modelling (MLM), makes self-supervised learning possible on unlabeled text sequences.
Afterward, it only requires fine-tuning on relatively small labeled datasets to usually get (near) state-of-the-art performance on a given language task, such as sentiment analysis, natural language inference and token tagging tasks.
However, such language models are difficult to deploy in production environments due to the fact that these models are large and thus require a lot of storage, and are slow and energy-intensive to perform inference on~\cite{bender2021parrots}.
Following the trend of distilling the knowledge from neural network models \cite{hinton2015distilling}, many types of distillation have been used to extract optimal parameters or extract the knowledge of larger language models into smaller ones \cite{sanh2019distilbert,dewynter2020bort,jiao2020tinybert}.
These smaller models require fewer resources and time to run, at the cost of being slightly less accurate.
Such a distillation thus allows for a favorable trade-off between performance and ease of use at deployment.

In this paper, we distill the Dutch BERT model RobBERT v2 \cite{delobelle2020robbert}, and name it RobBERTje\footnote{Dutch for ``Little RobBERT''}.
We perform several distillations using a small unlabeled Dutch dataset and fine-tune them on several language tasks to find the best processing of the dataset and target architecture hyperparametrizations.
The contributions of this paper are thus: (1) evaluating data processing for distillation;
(2) replicating studies on distillation
architectures; and (3) creating a more lightweight version of RobBERT to enable more efficient fine-tuning and energy-efficient inferencing of Dutch downstream language tasks.

\section{Background \& Related Work}

\subsection{BERT-like Models}

The BERT model is a powerful pre-trained language model that is used for training a vast number of more specific models for downstream natural language processing (NLP) tasks \cite{devlin2019bert}.
It is a bidirectional language model that is implemented using a transformer encoder stack, which exists of self-attention heads \cite{vaswani2017attention}.
By repeatedly applying these self-attention encoders, it is able to learn highly contextualized embeddings for each word token.
The insights learned during the pre-training phase have proven to be useful for many other linguistic tasks when researchers fine-tuned them on a wide range of other classification, regression and token tagging tasks, such as sentiment analysis, part-of-speech tagging and named entity recognition \cite{devlin2019bert}.
A BERT model is pre-trained with unlabeled data using the masked language modeling (MLM) task and the next sentence prediction (NSP) task.
The MLM task randomly masks tokens from a sentence and asks the BERT model to fill in the masked token.
The NSP task asks the BERT model to predict whether two sentences follow each other or are randomly sampled in the text.

The RoBERTa model replicated the BERT model and robustly optimized it while still following the same architecture as the BERT model \cite{liu2019roberta}.
It found that the NSP training task was redundant, and removed it from its pre-training regime.
The RoBERTa model also further improved the BERT architecture by changing its tokenizer to create a different vocabulary.
These optimizations increased its performance on most of the downstream NLP tasks.
In all other aspects, the RoBERTa model and the BERT model are usually mostly the same, and most findings on either model tend to also apply to the other.
An often used umbrella term for these types of models and other similar optimized BERT models is ``BERT-like models''.

Monolingual BERT-like models frequently outperform multilingual models, which are trained on many languages simultaneously, for most popular language tasks~\cite{nozza2020mask}
Therefore, a large number of BERT-like models have been trained using monolingual corpora.
A popular dataset containing these monolingual corpora for training monolingual BERT-like models is the OSCAR corpus.
This corpus is automatically constructed by using language classification on the web-crawled Common Crawl dataset.
Other researchers opt for building their own collection of training corpora, as this can allow them to perform better on tasks for their goal domain \cite{rasmy2021medbert,gu2021biomedbert}.
One monolingual Dutch BERT model is called BERTje, a model using the default BERT architecture trained on 2.4B tokens of selected formal Dutch text \cite{devries2019bertje}.
Another Dutch BERT-like model that was released around the same time is RobBERT, which uses the improved RoBERTa architecture and was trained on a larger corpus of 6.6B tokens of web text from the Dutch OSCAR dataset \cite{delobelle2020robbert}.
This improved architecture and larger training dataset allow it to outperform BERTje on most language tasks.
The RobBERT model also achieved state-of-the-art results on many Dutch NLP tasks compared to other types of models and has been used by numerous Dutch NLP researchers and practitioners since its release.
RobBERT forms a good basis for replicating several BERT distillation studies for Dutch and also allows us to investigate several properties of distilled models, for example by altering its distillation dataset.

\subsection{Knowledge Distillation}

Knowledge distillation is the technique used for learning a simpler model (student) from a more complex model (teacher).
Initially, this technique was called model compression and used a large ensemble model (as a  teacher) to label a large unlabeled dataset for the student model to learn from \cite{bucilua2006modelcompression}.
The main advantage thus came from the student having access to a larger dataset, even if this leads to somewhat noisy labels due to mistakes made by the teacher model.
This student model can then be used instead of the teacher model in certain situations thanks to being smaller and thus faster and less resource-intensive at the cost of lower accuracy.

The model compression technique was later further extended for neural networks in a process called knowledge distillation.
This distillation uses the fact that neural networks typically predict probabilities for each possible label by producing class probabilities using a softmax output layer.
In a neural network setting the student can then learn from the probabilities assigned by the teacher to the incorrect labels, thus learning to generalize the same way the teacher model does \cite{hinton2015distilling}.
These label probability distributions $z_i$ (also called soft targets) are estimated using the softmax function in Equation \ref{eq:softmax_temperature}, where $T$ is a temperature controlling the soft target importance, as higher values produce softer probability distributions \cite{hinton2015distilling}. 
This temperature $T$ also acts as a regularizer during training \cite{hinton2015distilling}.

\begin{equation}
\label{eq:softmax_temperature}
p(z_i, T) = \frac{e^{z_i/T}}{\sum_j e^{z_j/T}}
\end{equation}

The distillation algorithm then trains the student model using a dataset as a transfer dataset, predicting the probabilities for each label using both the student and the teacher, and using cross-entropy as loss function $L_{ce}$ between these predictions for the same data point (see Equation \ref{eq:student_teacher_loss}, where $t_i$ and $s_i$ are the predictions by the teacher and student respectively).
This way, the student learns to approximate the predictions for all labels from its teacher.

\begin{equation}
\label{eq:student_teacher_loss}
L_{ce} = \sum_ip(t_i, T) * log(p(s_i, T))
\end{equation}

While this method was initially introduced to compress ensemble models into simple neural networks, it has been used for a wide variety of other similar distillations.
For example, it was later used to also distill neural networks to similar networks with fewer layers and neurons or with more efficient basic operators, and it has been suggested as a means for discovering good student architectures \cite{gou2021knowledgedistillationsurvey}.

\subsection{BERT Distillation}\label{ss:distillation-background}

With the rise of large-scale pre-trained language models with hundreds of millions of parameters like BERT-like models, there is an alarming trend towards bigger models to get even higher accuracy on downstream tasks \cite{sanh2019distilbert}.
As these models are scaling exponentially, distilling such large language models has received a lot of attention.
BERT-like models have been distilled using a wide range of distillation methods in order to make them more suitable for real-world applications.
These distilled BERT-like models are easier to deploy, less resource-intensive to train and/or less time-consuming to perform inference on.
The goal of these distillation is usually to make a much smaller model without having to sacrifice much accuracy on the target downstream task.

BERT-like models have two different training phases, namely a large, general pre-training phase and a small, specific fine-tuning phase.
Model distillation can happen after either training phase.
Depending on whether the model is distilled after pre-training or after fine-tuning, the distilled model functions either still as a BERT-like model (i.e. general BERT distillation), or just as a model for this particular task (i.e. task-specific BERT distillation).
Sometimes these approaches are mixed in a two-stage distillation model, such as TinyBERT, which first performs general transformer distillation, and then fine-tunes the model via task-specific distillation \cite{jiao2020tinybert}.

\subsubsection{Task-Specific BERT distillation}

In task-specific BERT distillation, a large BERT-like model is fine-tuned for a particular task, and then afterward distilled into a much smaller student network that can then only perform this specific downstream task.
This smaller student network is often a completely different type of neural network architecture than BERT, e.g. an LSTM-based classifier.
It has been shown that fine-tuned BERT-like models can be distilled to a BiLSTM with the number of parameters cut to 1/100 and inference time to 1/15 of the original model's values
and still achieve comparable results on language tasks such as paraphrasing, natural language inference and sentiment classification \cite{tang2019distillingtaskspecific}.

\subsubsection{General BERT distillation}
General BERT distillation distills a pre-trained BERT-like model and aims to still retain the same properties as the original pre-trained model.
The resulting student model is usually a similarly structured but smaller BERT-like architecture and thus can then still be fine-tuned for other downstream tasks, just like its teacher \cite{sanh2019distilbert}.
One of the reasons why distilled general BERT models still have similar accuracy on downstream tasks is because the BERT model is significantly overparametrized \cite{kovaleva2019revealingbert}.
Most heads in the same layer contain self-similar attention patterns \cite{clark2019does}.
Due to BERT containing a lot of redundant heads given the rest of the model, 20\% to 40\% of the heads can be pruned without noticeable negative impact  \cite{michel2019sixteen}.
In fact, disabling attention in certain heads of the BERT model can even lead to performance improvement \cite{kovaleva2019revealingbert}.

DistilBERT employs knowledge distillation by learning the probability distribution for tokens in the MLM task.
Its student model is created by removing the token-type embeddings and pooler (for the NSP task) from BERT, and halving the number of layers, while the rest of the BERT architecture is kept identical.
The distillation uses three loss functions, namely the $L_{CE}$ (Equation~\ref{eq:student_teacher_loss}), the masked language modeling loss $L_{mlm}$ \cite{devlin2019bert} and cosine embeddings loss $L_{cos}$ \cite{sanh2019distilbert} to align student and teacher hidden state vector directions.
The resulting model had 40\% fewer parameters and still retained 97\% its language understanding while being 60\% faster \cite{sanh2019distilbert}.

Determining the size and architecture of the student model is a non-trivial task.
Researchers found optimal sizes for the student architecture, both experimentally \cite{turc2019well} and using formal optimal parameter extraction methods such as Bort \cite{dewynter2020bort}.
Bort was proposed as an optimal subset of BERT's architectural parameters, and it is architecturally similar to BERT and uses the RoBERTa tokenizer.
Instead of deciding the architecture parameters arbitrarily, the authors of Bort attempted to discover optimal parametrizations.
These Pareto optimal architectural parameters are supposed to balance the inference speed, parameter size and error rate.
The Bort parameters specify a model that is 16\% the size of its BERT-large teacher and 20 times faster than BERT-large on a CPU on a wide range of language tasks \cite{dewynter2020bort}.
While the found Bort hyperparametrizations are known (D=4, A=8, H=1024, I=768), its finetuning algorithm Agora is not publicly available.

\section{RobBERTje Distillation Experiments}
\label{sec:choices}

There are several types of choices when performing distillation on a BERT-like model.
While previous research has explored several distillation algorithms and architectural parameters, little studies have evaluated the importance of the transfer dataset.
For example, there has been some disagreement whether training on the non-shuffled or shuffled versions of the OSCAR training dataset influences the performance of a pre-trained BERT-like model positively or negatively \cite{wouts2020belabbert,delobelle2020robbert}.
The same question can be raised about the transfer dataset in BERT distillation.
Similarly, no research has tried to replicate the Bort research by using its found optimal hyperparameters for a language other than English.

To perform our distillation experiments, we distill several smaller models from the Dutch RobBERT model.
Since this model achieves state-of-the-art results on many downstream Dutch language tasks~\cite{delobelle2020robbert}, distilling these models allows us to not only evaluate our hypotheses but also provide the Dutch NLP community with smaller, near state-of-the-art Dutch language models.
As RobBERT uses the RoBERTa architecture~\cite{liu2019roberta} and the OSCAR corpus~\cite{ortizsuarez2019oscar}, we decided that the distilled RobBERT models should also use smaller versions of both for its architecture and transfer dataset.
We then experimented with the influence of order and the length of the transfer dataset and replicated studies of the DistilBERT and Bort architectures.

As there are many choices to make when distilling a model, we test out several choices for the data and target distillation architecture.
More specifically, we test whether it matters if the training corpus is shuffled (\autoref{ss:shuffle}), the influence of the length of the training sequences (\autoref{ss:merge}) and what distillation architecture hyperparametrization works best for the distilled model (\autoref{ss:bort}).

\subsection{To Shuffle Or Not To Shuffle?}
\label{ss:shuffle}

The OSCAR 2019 corpus~\cite{ortizsuarez2019oscar} is one of the most used datasets to train large language models and is publicly available in a shuffled form for obfuscation purposes.
It is constructed by automatically classifying the language of the web-crawled CommonCrawl dataset. 
The original, non-shuffled variant is also available upon request.
While some hypothesized that using a non-shuffled version could allow the model to learn dependencies spanning multiple sequences~\cite{wouts2020belabbert}, the order itself might also not be important for pre-training because each input sequence is used individually. 
Since RoBERTa dropped next-sentence prediction due to it being an ineffective pre-training task, models using this optimized training regime also lack these longer connections across separate training sequences~\cite{liu2019roberta}.
It is possible that not shuffling the dataset could hurt the training performance due to less diverse training sequences in every batch.
These batches could then potentially be less representative of the true gradient over the whole dataset, thus pushing the gradient into less desirable directions.
To get more insights on the advantages and disadvantages of shuffling the transfer datasets, we set up an experiment where we distilled two models (\emph{Shuffled} and \emph{Non-shuffled}) using the DistilBERT regime, where only the nature of the transfer dataset was different.

\subsection{Sequence Merging for Increased Sequence Length}
\label{ss:merge}

\begin{figure}[t]
    \centering
    \includegraphics[width=0.9\textwidth]{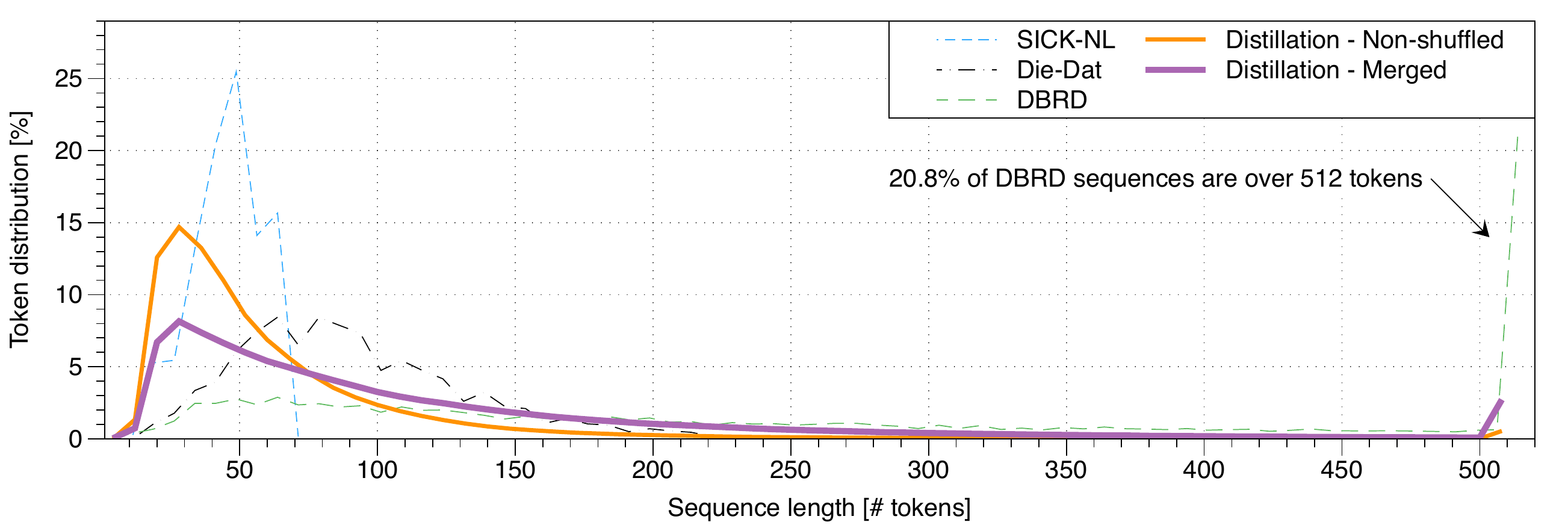}
    \caption{Distribution of the sequence length of non-shuffled OSCAR with and without merging sentences, and the sequence lengths of several task datasets.}
    \label{fig:oscar-distribution}
\end{figure}

Another unexplored question is the impact of the length of the unlabeled text sequences of the transfer dataset on the performance of the resulting distilled models.
For example, the Dutch OSCAR corpus has mostly relatively short sequences ($<40$ tokens, \autoref{fig:oscar-distribution}), which may or may not comprise multiple sentences.
However, since the OSCAR corpus marks the start and end of the documents, these short sequences arise from using newlines and it is possible to concatenate these related text sequences into valid longer ones.
We thus derive a new transfer dataset by concatenating two sequential lines from the same document into one training sequence with a probability $p=0.5$.
We hypothesize that using longer sequences allows later input positions of the distilled model to see more actual data instead of padding tokens.
This is important, as up to 512 input token positions are uniquely encoded with a positional encoding.
Given the fact that OSCAR mostly has short sentences, later positions do not have as many training examples as earlier positions, which might affect tasks containing important information at the end of the input positions.
Merging subsequent sequences from the same documents could thus theoretically improve the performance of downstream tasks that involve processing long sequences.
An additional benefit is that merging sequences compacts the dataset into fewer sequences, thus decreasing training time and energy for pre-training and distillation.
Since a BERT-like model always processes all input tokens anyway, the longer lengths do not influence training time.
One downside is that this leaves relatively less training data for the initial input positions of the model compared to the original version using a non-merged corpus (e.g. \emph{Non-shuffled}).

\begin{figure}[h]
\centering
    \includegraphics[width=0.7\textwidth]{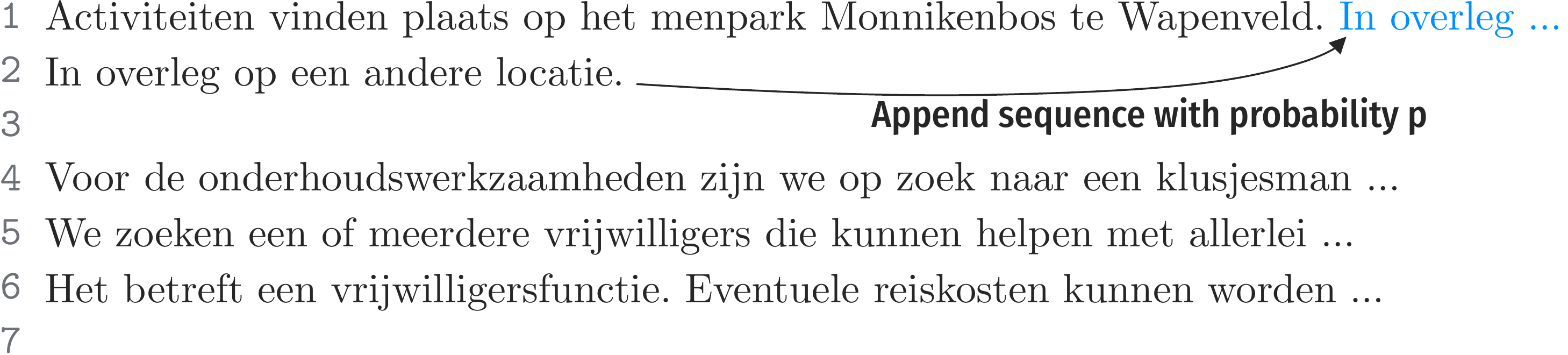}
    \caption{Excerpt from the Dutch section of OSCAR to highlight how different sequences, stored and represented as different lines, are merged within a document.}
    \label{fig:example-sentence}
\end{figure}

We created a new dataset from the non-shuffled Dutch OSCAR dataset by randomly merging a sequence with its following sequence with a probability of 50\% if they are from the same document.
This resulted in a smaller corpus with generally longer sequences (\autoref{fig:oscar-distribution}), thus reducing the time required to perform the distillation.
After the sequence merging, we shuffled the sequences from the resulting, merged corpus before using it as a transfer dataset.

\subsection{Target Architecture}
\label{ss:bort}

There are several choices when it comes to choosing the student architecture and its hyperparametrization for distillation, as we discussed in \autoref{ss:distillation-background}.
For the earlier two experiments, we used the DistilBERT architecture hyperparametrizations \cite{sanh2019distilbert}, which roughly halves the number of parameters and keeps the general properties of the teacher (RobBERT).
As mentioned earlier, the Bort model recently emerged with a claim to have found an optimal hyperparametrization, balancing inference speed, parameter size and error rate \cite{dewynter2020bort}.
The Bort model also uses only 56M parameters, while its teacher BERT-large uses 340M (for reference, the BERT-base model uses 110M parameters).
We aim to replicate this study using the same student architecture with the same hyperparameters and test whether these parametrizations are still optimal when used on this Dutch RobBERT model.
However, we kept the hidden size to 768 to allow for hidden distillation, which is the same hidden size as the teacher model.
We used the \emph{Merged} subsequences transfer dataset to distill this Dutch Bort model from its RobBERT teacher.
Since the specialized code used for fine-tuning in the Bort paper is not publicly available, we use the same fine-tuning procedures as \citeasnoun{sanh2019distilbert} to fine-tune the resulting distilled model.

\section{Experimental setup}
In this section, we provide an overview of the Dutch language tasks and the corresponding datasets we used to evaluate our models.
We also give a detailed overview of our distillation and fine-tuning setup.

\subsection{Benchmark Task Datasets}
\label{ss:benchmarks}

We evaluated the performance of the distilled models described in the previous section on six types of language tasks.
This wide range of downstream language tasks allows us to gain insights into which models perform better on what types of tasks.
These tasks overlap with evaluation tasks in the original RobBERT paper, which are described in more detail in that paper \cite{delobelle2020robbert}.

\subsubsection{Sentiment Analysis (SA)}
Sentiment analysis is a language task in which the model needs to predict subjective information of a text, e.g. whether a given article has a more positive or more negative sentiment.
To evaluate the sentiment analysis performance, we use the Dutch Book Reviews Dataset (DBRD) \cite{vanderburgh2019dbrd}, which is a binary classification sentiment analysis dataset.
It contains 22K book reviews with a label denoting whether the review was a positive (4-5 stars on the \emph{Hebban.nl} book reviews website) or a negative review (1-2 stars).
The text sequences are generally rather long, with 20.8\% of the reviews longer than the maximum input size of the RobBERT and RobBERTje models (Figure \ref{fig:oscar-distribution}).
For these long reviews, we use the last 512 tokens of the sequence, as these were also used instead of the first 512 tokens as in \citeasnoun{delobelle2020robbert}, which found that this choice leads to better performance in the RobBERT evaluation.

\subsubsection{Co-reference Resolution (CR)}
Co-reference resolution is a language task in which the model predicts which parts of a sentence reference the same entity, e.g. matching pronouns with the named entity earlier or later in a sentence.
To evaluate the performance of the distilled models on this type of task, we make the models predict whether a sentence needs to be filled with \emph{die} or \emph{dat} on a given position, as the choice depends on which word it refers to.
We use the EuroParl dataset \cite{koehn2005europarl}, which contains the proceedings from the European Parliament, to predict the pronouns in.
This corpus was also used when this \emph{die-dat}-disambiguation task was initially introduced by \citeasnoun{allein2020diedat}.
In the dataset, 947k training sentences are used for training the model to predict whether ``die'' or ``dat'' should be inserted in a position in the sentence, 237k sentences are used as validation and 305k for testing.

\subsubsection{Named Entity Recognition (NER)}
In a named entity recognition task, a model needs to predict which parts of a text sequence are named entities, and often also what type of named entity.
In our experiments, we evaluate the named entity recognition capabilities by using the CoNLL-2002 dataset \cite{sang2002conll}.
In this dataset, words from sentences are tagged as starting or continuing a named entity, and also what type of entity (person, organization, location or miscellaneous).
The training data contains 15.8K sequences, with 2.8K validation sequences and 5.1K test sequences.

\subsubsection{Part-of-Speech Tagging (POS)}
Part-of-speech tagging is a language task in which a model predicts the part-of-speech tag (e.g. adjective, noun, etc) for each word of a text sequence.
For our experiments, we used the universal dependencies version of the Lassy dataset \cite{vannoord2013lassy}.
This dataset contains sentences where each word is tagged as the beginning or continuing a certain part-of-speech tag and contains 5787 training examples, 676 examples for validation and 875 examples for testing.

\subsubsection{Natural Language Inference (NLI)}

Natural language inference is a language task in which a model needs to predict for two statements whether the second statement is a consequence, contradiction or neither of the first statement.
We use SICK-NL \cite{wijnholds2021sicknl} as the natural language inference dataset for our experiments.
This dataset is a semi-automatically translated version of the original SICK natural language inference dataset \cite{marelli2014sick}, which contains sentence pairs annotated with their relatedness (score from 1 to 5) and entailment (entailment, contradiction and neutral).
We modified the SICK-NL dataset by adding a period at the end of the sentences, as this significantly improves the performance for BERT models due to the fact that these models are generally trained on well-formed sentences with complete punctuation.

\subsubsection{Pseudo-Perplexity (PPPL)}

Perplexity is a metric for evaluating language models and is defined as the exponentiated average negative log-likelihood of a sequence, thus indicating how well a model can predict the right token.
Since BERT models are generally not well suited for this metric, an alternative pseudo-perplexity (PPPL) was proposed for measuring how well the MLM models a corpus of sentences \cite{salazar2019mlmscoring}.
For each input sentence, the PPPL algorithm creates all possible versions of this sentence with one masked token and then multiplies the probabilities for all sentences.
We used the last segment of the non-shuffled OSCAR corpus as the evaluation data.
While our previous language tasks all measure the performance of fine-tuned versions of the distilled models, the PPPL metric allows us to measure the MLM quality of the distilled model itself\footnote{We used a script at \url{https://github.com/iPieter/universal-distillation}.}.

\subsection{Setup}
After distillation, we fine-tune all 4 distilled model variants on the trainable tasks (SA, CR, NER, POS, and NLI) discussed in \autoref{ss:benchmarks}.
For each of the 5 fine-tuning tasks, we train 5 models with random hyperparameters (a full list is provided in \autoref{tab:hp-space}), resulting in 100 fine-tuned models in total. 
We select the best-performing model on the validation set and evaluate this on the test set, of which the results are reported.

All fine-tuned models were trained on 1 Nvidia 1080 Ti GPU with a batch size of 8.  
Because the distillation objective required loading both the teacher model and the smaller trainable model, the batch size was slightly lower, namely 5. 
To improve stability during distillation, we accumulated gradients for 128 steps, giving an effective batch size of 640.
We perform these experiments using the first 1GB of the non-shuffled Dutch OSCAR dataset using one Nvidia 1080 Ti.
For the MLM perplexity evaluation, we use 50k sequences from the last shard of the non-shuffled dataset.

To aid reproducibility and to promote further fine-tuning on these smaller but effective language models, we release our distilled RobBERT models, as RobBERTje, as well as the training configurations, on \url{https://github.com/ipieter/robbertje} and on HuggingFace's Hub under the ``\texttt{DTAI-KULeuven/robbertje-}'' prefix.

\section{Results}

We present the results of our experiments in \autoref{tab:results}, where we also include some hyperparameters and the size of the training corpora.
For comparison, we also included another Dutch BERT model called BERTje~\cite{devries2019bertje}.
We discuss the results of each experiment separately. 

\subsection{Shuffled versus Non-Shuffled}
\label{ss:results-shuffle}

In the first experiment, we tested the influence of shuffling the transfer dataset on the performance of the resulting distilled model.
This is achieved by distilling two RobBERTje models, which only differ in the fact that \emph{Shuffled} uses the shuffled OSCAR corpus, and \emph{Non-shuffled} the original OSCAR corpus.
The differences between these two models are very small for most downstream language tasks.
Not shuffling the training data appears to give rise to a better MLM head in the distilled pre-trained model, as its pseudo-perplexity (PPPL) is much lower and thus better.
Thee \emph{Shuffled} model performs much better on sentiment analysis.

Compared to both RobBERT and BERTje, we observe that the performance trade-off varies between tasks. 
On most tasks, like NLI and POS tagging, there is only a slight decrease compared to the teacher model.
Interestingly, the distilled \emph{Non-shuffled} model even performs slightly better than the larger BERTje on co-reference resolution.
The performance of both \emph{Non-shuffled} and \emph{Shuffled} on the NER and PPPL tasks are in contract much lower compared to their teacher RobBERT.

\subsection{Effect of Sequence Merging}

We distilled a model called \emph{Merged} that only differs with the \emph{Shuffled} model in that it first concatenates some subsequent sequences of the transfer dataset (as explained in \autoref{ss:merge}).
We hypothesized that merging data into longer subsequences is advantageous to tasks using long sequences as input, as the later input tokens see relatively more input.
We also hypothesized that merging subsequent sequences would likely be detrimental for tasks that deal with shorter sequences as they processed fewer training sequences than without merging.
We see that the \emph{Merged} model acts according to our hypotheses compared to its non-merged counterparts.
It performs better than both on the sentiment analysis task, which uses the long movie reviews of the DBRD dataset that often use all input tokens (Figure \ref{fig:oscar-distribution}).
Similarly, it performs worse on tasks that have shorter sequences such as SICK-NL, which has the shortest sequences of all tasks (as can be seen on \autoref{fig:oscar-distribution}). 
Similarly, there is another large trade-off for co-reference resolution, which also uses more of the early input tokens.
Thus, we recommend this model only for downstream tasks that require the full input token length.

\subsection{Target Architecture Hyperparametrization}

We evaluated the performance when using the Bort hyperparametrizations on the Dutch RobBERT model by distilling a model with these architecture sizes.
Our distilled \emph{Bort} model is much smaller and faster than the other DistilBERT-based models. For example, fine-tuning to SICK-NL is 4 times faster than RobBERT and 2.2 times faster than our merged sequence distillation.
However, while it might be much smaller, it is significantly outperformed by its DistilBERT counterparts on all tasks.
It also has a much worse performance on the tasks compared to its teacher than one might expect from the results of the original English Bort \cite{dewynter2020bort}.
As this set of hyperparameters was found to be optimal for the English RoBERTa model~\cite{dewynter2020bort}, these results are quite surprising.
One possible explanation is that we had to use default fine-tuning algorithms instead of the specialized Bort fine-tuning algorithm called Agora because this algorithm was not made public.
Regardless, we were thus unable to find evidence that their calculated optimal hyperparameters also work for Dutch BERT models.

\begin{table}[t]
\centering
\caption{Overview of all pretrained models and benchmark results and number of decoder layers $D$, number of attention heads $A$, hidden size $H$ and intermediate layer size $I$. We report accuracy and 95\% CI for all benchmark scores, except NER, which uses the $F_1$ score. PPPL results for RobBERT are indicative, since we could not guarantee the pre-training data was not seen before. \emph{(Results indicated with $^{*}$ are reported their respective authors.})}\label{tab:results}
\resizebox{\textwidth}{!}{%
\begin{tabular}{@{}lrrrrrrlllllr@{}}
\toprule
              & \multicolumn{1}{l}{}                 & \multicolumn{4}{c}{\textsc {\textbf{Hyperparameters}}}                                                        & \multicolumn{1}{l}{\textbf{}}       & \multicolumn{5}{c}{\textsc {\textbf{Benchmark scores}}}                                                                                        \\ \cmidrule(lr){3-6} \cmidrule(l){8-13} 
\textbf{Model}& \multicolumn{1}{l}{\textbf{Data}} & \multicolumn{1}{c}{\textbf{D}} & \multicolumn{1}{c}{\textbf{A}} & \multicolumn{1}{c}{\textbf{H}} & \multicolumn{1}{c}{\textbf{I}} & \multicolumn{1}{c}{\textbf{Params}} & \multicolumn{1}{c}{\textbf{SA}} & \multicolumn{1}{c}{\textbf{CR}} & \multicolumn{1}{c}{\textbf{NER}} & \multicolumn{1}{c}{\textbf{POS}}  & \multicolumn{1}{c}{\textbf{NLI}}& \textbf{PPPL}\\ \midrule
RobBERT v2       & 39 GB            & 12         & 12         & 768        & 3072       & 116 M           & $ 94.4 \pm 1.0^{*}$                   & $99.2 \pm 0.03^{*}$             & $89.1^{*}$             & $96.4\pm 0.4^{*}$ & $84.2 \pm 1.0$                            & \textit{7.76}   \\ 
BERTje       & 12 GB             & 12         & 12         & 768        & 3072       &  109~M           & 93.0$^{*}$                  &    98.3$^{*}$          & 88.3$^{*}$            & 96.3$^{*}$ & 83.94$^{*}$                          & 12.22   \\ \hline

Non-shuffled (\autoref{ss:shuffle})  & 1 GB             & 6          & 12         & 768        & 3072       & 74 M            & $90.2 \pm 1.2$& $\mathbf{98.4} \pm 0.1$                    &  $\mathbf{82.9}$            & $95.5 \pm 0.4$ & $\mathbf{83.4} \pm 1.0$   & \textbf{12.95} \\
Shuffled (\autoref{ss:shuffle})      & 1 GB             & 6          & 12         & 768        & 3072       & 74 M            & $92.5 \pm 1.1$& $98.2 \pm 0.1$                    & $82.7$             &$\mathbf{95.6} \pm 0.4$ & $\mathbf{83.4} \pm 1.0$                           & 18.74       \\
Merged (\autoref{ss:merge})          & 1 GB             & 6          & 12         & 768        & 3072       & 74 M            & $\mathbf{92.9} \pm 1.1$& $96.5 \pm 0.1$     &  $81.8$           & $95.2 \pm 0.4$      &  $82.8 \pm 1.1$                        & 17.10        \\
Bort (\autoref{ss:bort})             & 1 GB             & 4          & 8          & 768        & 768        & 46 M            & $89.6 \pm 1.3$& $92.2 \pm 0.1$                    & $79.7$             & $94.3 \pm 0.4$ &  $81.0 \pm 1.1$                     & 26.44           \\
  \bottomrule

\end{tabular}%
}
\end{table}

%D = encoder layers, 
%A = attention heads, 
%H = hidden size, and 
%I = intermediate layer size

\section{Limitations and fairness}
\citeasnoun{delobelle2020robbert} also presented an in-depth fairness analysis of their model, investigating both intrinsic and extrinsic forms~\cite{blodgett2020language} of gender bias. 
Because stereotypes, biased language and even hate speech all occur in the datasets on which many language models are trained, including OSCAR for RobBERT and RobBERTje~\cite{caswell2021quality}, these models are capable of replicating these input patterns.
This results in many observed problematic correlations~\cite{may2019,blodgett2020language,webster2020,delobelle2020robbert}.
This sparked the creation of metrics to quantify these correlations in LMs like BERT, some based on previous works on bias in word embeddings~\cite{bolukbasi2016man}.
For an overview and comparison of such metrics, we refer the reader to \citeasnoun{delobelle2021measuring}.

The \emph{Word Embedding Association Test} (WEAT) \cite{caliskan2017weat} is one such metric that was later extended to the Sentence Embedding Association Test (SEAT) using templates~\cite{may2019}.
\citeasnoun{kurita2019measuring} observe that using WEAT for the learned BERT embeddings fails to find many statistically significant biases, which is addressed in the presented \emph{log probability bias score}. 
This score computes a probability $p_{tgt}$ for a target token~$t$ (e.g. `He' or `She') from the distribution of the masked position $X_m$ following
\[ 
  p_{tgt} = P\left(X_m = t \mid \mathbf{x}; \theta \right),
\]
for a template sentence, e.g. ``{\tt <mask>} is a nurse'' with {\tt <mask>} indicating the masked position $X_m$.
Since the prior likelihood $P(X_m=t)$ can skew the results, the authors correct for this by calculating a template prior $p_{prior}$ by additionally masking the token(s) with a profession or another attribute $x_p$, following

\[ 
  p_{prior} = P\left(X_m = t \mid \mathbf{x} \backslash \{x_p\}; \theta \right).
\]
Both probabilities are combined in a measure of association $\log\frac{p_{tgt}}{p_{prior}}$ and the bias score is the difference between the association measures for two targets, like `He' and `She'.
\citeasnoun{kurita2019measuring} applied their method on the original English BERT model~\cite{devlin2019bert} and found statistically significant differences for all categories of the WEAT templates. 
We use this metric in combination with the translated list of professions by \citeasnoun{delobelle2020robbert} to evaluate gender stereotyping in our distilled models, as shown in \autoref{tab:bias}.

\begin{table}[]
\centering
\caption{Log probability bias score for RobBERT and our RobBERTje models. Positive scores indicate higher correlations with gender stereotypes of professions.}\label{tab:bias}
\begin{tabular}{@{}ll@{}}
\toprule
                                   & \textbf{Bias score} \\ \midrule
Teacher \cite{delobelle2020robbert} & 1.10       \\\midrule
Non-shuffled (\autoref{ss:shuffle}) & -0.52      \\
Shuffled (\autoref{ss:shuffle})     & -0.50      \\
Merged (\autoref{ss:merge})         & -0.67      \\
Bort (\autoref{ss:bort})            & 0.04       \\ \bottomrule
\end{tabular}
\end{table}

The bias evaluations in \autoref{tab:bias} do show that the original RobBERT model was exhibiting some gender stereotyping with regards to professions, as noted before by \citeasnoun{delobelle2020robbert}.
The distilled models do seem to correct this stereotyping and all except the Bort model even overcompensate. 
\citeasnoun{webster2020} noted that regularization methods, in their case dropout, are effective in attenuating stereotypes. 
Since knowledge distillation with soft targets can be considered a form of regularization~\cite{hinton2015distilling}, we suspected that our distilled models would show a decrease in stereotypes.
The results in \autoref{tab:bias} confirm this.

\section{Future Work}

In this paper, we focused on general BERT distillation to create distilled versions of RobBERT that can still be fine-tuned in the same way.
It would be interesting to perform task-specific distillation or even two-stage distillation like TinyBERT on RobBERT and compare the performance against the general RobBERTje models.
Since models distilled from task-specific distillation do not need to be BERT-like models, they can be orders of magnitudes smaller and possibly more accurate on the target task.
In this paper, however, we focused on general BERT distillation as we believe this lowers the threshold of fine-tuning Dutch BERT-like models thanks to the lower computational requirements and faster inference times.
We hope that the Dutch NLP community can benefit from these models by fine-tuning a suitable RobBERTje model for their own downstream task with significantly less computing power and storage.

\section{Summary and Conclusions}

In this paper, we created multiple distilled versions of the state-of-the-art Dutch RobBERT model and called this family of models ``RobBERTje''.
In doing so, we found that the influence of using a shuffled dataset is small for distillation.
We also found that randomly merging subsequent sequences of the non-shuffled dataset improves the performance of the distilled language model for tasks using longer input sentences.
We replicated the Bort approach and found that while the model is much smaller than its DistilBERT counterpart, its performance was significantly worse on all tested tasks.
Interestingly, we found that in distilling the RobBERTje models, they all show less stereotypical bias than their teacher RobBERT due to the soft labels acting as a regularizer.
The overall results suggest that our new distilled RobBERTje models can be used for making a large number of downstream Dutch natural language processing tasks much more efficient while still achieving close to state-of-the-art results.

\subsection*{Acknowledgements}

We thank the anonymous reviewers for their valuable feedback.
Pieter Delobelle was supported by the Research Foundation - Flanders (FWO) under EOS No. 30992574 (VeriLearn).
Pieter Delobelle also received funding from the Flemish Government under the ``Onderzoeksprogramma Artificiële Intelligentie (AI) Vlaanderen'' programme.
Thomas Winters is supported by the Research Foundation-Flanders (FWO-Vlaanderen, 11C7720N). 
Bettina Berendt received funding from the German Federal Ministry of Education and Research (BMBF) – Nr. 16DII113f.

%Most computational resources and services used in this work were provided by the VSC (Flemish Supercomputer Center), funded by the Research Foundation - Flanders (FWO) and the Flemish Government – department EWI.

\bibliographystyle{clin} 
\bibliography{bibliography}  

\newpage

\appendix
\section{Hyperparameter space}

\begin{table}[h]
\centering
\caption{The hyperparameter space used for fine-tuning.}
\label{tab:hp-space}
\begin{tabular}{@{}ll@{}}
\toprule
\textbf{Hyperparameter}         & \textbf{Value}         \\ \midrule
adam\_epsilon                   & $10^{-8}$              \\
fp16                            & False                  \\
%fp16\_opt\_level                & O1                     \\
gradient\_accumulation\_steps   & $i \in \{2, 4, 8, 16\}$ \\
learning\_rate                  & $[10^{-6}, 10^{-4}]$   \\
max\_grad\_norm                 & 1.0                    \\
max\_steps                      & -1                     \\
num\_train\_epochs              & 3                      \\
per\_device\_eval\_batch\_size  & 8                      \\
per\_device\_train\_batch\_size & 8                      \\
max\_sequence\_length.          & 512 \\
seed                            & 1                      \\
warmup\_steps                   & 0                      \\
weight\_decay                   & $[0, 0.1]$             \\ \bottomrule
\end{tabular}
\end{table}

\end{document}